\documentclass[letterpaper, 10 pt, journal, twoside]{IEEEtran}
\usepackage[table]{xcolor}

\makeatletter
\let\NAT@parse\undefined
\makeatother
\usepackage[numbers, sort&compress]{natbib}

\usepackage[T1]{fontenc}
\usepackage{lmodern}

\usepackage{times}

\usepackage{tikz}
\usepackage{graphicx} 
\usepackage{amsmath} 
\usepackage{stfloats}
\usepackage{makecell}
\usepackage{arydshln}
\usepackage{bigstrut}

\usepackage{amsmath}
\usepackage{amssymb}  
\usepackage{amsfonts}
\usepackage{subfigure}
\usepackage{ctable}
\usepackage{url}
\usepackage{makecell}

\usepackage[titlenumbered,linesnumbered,ruled,vlined]{algorithm2e}

\usepackage{fancyvrb}
\usepackage{xcolor}
\usepackage{verbatimbox}

\usepackage[labelsep=period]{caption}
\usepackage[numbers]{natbib}
\usepackage{multicol}
\usepackage[bookmarks=true]{hyperref}

\usepackage{amsthm}
\theoremstyle{definition}

\usepackage{tool_misc}
\usepackage{url_acronyms}

\usepackage{soul,color}
\usepackage{verbatim} 
\usepackage{multirow}

\usepackage{hyperref}
\usepackage{textcomp}
\usepackage{lipsum,multicol}
\usepackage{xcolor}

\DeclareMathOperator*{\argmin}{argmin}

\begin{document}
	
	\definecolor{revised}{rgb}{0, 0, 1.0}
	
	\title{\LARGE \bf
		DynaVINS++: Robust Visual-Inertial State Estimator in Dynamic Environments
		by Adaptive Truncated Least Squares and Stable State Recovery
	}
	
	\author{Seungwon Song${}^{1}$, Hyungtae Lim${}^{2}$, Alex Junho Lee${}^{1}$, and Hyun Myung${}^{3*}$,~\IEEEmembership{Senior~Member, IEEE} 
	\vspace{-0.8cm}
		\thanks{Manuscript received: April 27, 2024; Revised: July 24, 2024; Accepted: August 21, 2024. 
			This paper was recommended for publication by Senior Editor Pascal Vasseur upon evaluation of the Associate Editor and Reviewers' comments.} 
		\thanks{$^{1}$Seungwon Song and $^{1}$Alex Junho Lee are with the Robotics Lab, Research and Development Division, Hyundai Motor Company, Uiwang 16082, South Korea. {\tt\footnotesize \{ssw.robotics,alexjunholee\}@hyundai.com}}
		\thanks{$^{2}$Hyungtae Lim is with the Laboratory for Information \& Decision Systems, Massachusetts Institute of Technology, Cambridge, MA 02139, USA.  {\tt\footnotesize shapelim@mit.edu}}
		\thanks{$^{3}$Hyun Myung is with the School of Electrical Engineering, KAIST (Korea Advanced Institute of Science and Technology), Daejeon, 34141, Republic of Korea. \tt\footnotesize hmyung@kaist.ac.kr}
		\thanks{This work was supported by Korea Research Institute for defense Technology planning and advancement(KRIT) grant funded by the Korea government(DAPA(Defense Acquisition Program Administration)) (No. KRIT-CT-22-006(04), Development of micro-swarm robot autonomous navigation technology, 2022), and the Unmanned Swarm CPS Research Laboratory program of Defense Acquisition Program Administration and Agency for Defense Development(UD220005VD).}
		\thanks{Dr. Hyungtae Lim was supported by the Basic Science Research Program through the National Research Foundation of Korea (NRF) funded by the Ministry of Education (RS-2024-00415018) and BK21 FOUR.}
		\thanks{* Corresponding author: Prof. Hyun Myung}
	}
	\markboth{IEEE Robotics and Automation Letters, 06 September 2024. DOI: 10.1109/LRA.2024.3455905}
	{Song \MakeLowercase{\textit{et al.}}: DynaVINS++: Robust Visual-Inertial State Estimator in Dynamic Environments} 
	
	\maketitle
	\begin{abstract}
Despite extensive research in robust visual-inertial navigation systems~(VINS) in dynamic environments, many approaches remain vulnerable to objects that suddenly start moving, which are referred to as \textit{abruptly dynamic objects}.
In addition, most approaches have considered the effect of dynamic objects only at the feature association level.
In this study, we observed that the state estimation diverges when errors from false correspondences owing to moving objects incorrectly propagate into the IMU bias terms.
To overcome these problems, we propose a robust VINS framework called \mbox{\textit{DynaVINS++}}, which employs
a) adaptive truncated least square method that adaptively adjusts the truncation range using both feature association and IMU preintegration to effectively minimize the effect of the dynamic objects while reducing the computational cost,
and b)~stable state recovery with bias consistency check to correct misestimated IMU bias and to prevent the divergence caused by abruptly dynamic objects.
As verified in both public and real-world datasets,
our approach shows promising performance in dynamic environments, including scenes with abruptly dynamic objects.
\end{abstract}
	

\section{Introduction}
\label{sec:intro}

\IEEEPARstart{V}{isual}-inertial odometry~(VIO) is a methodology that utilizes both images from a camera and data from inertial measurement unit~(IMU) to tackle the scale ambiguity problem in visual odometry~(VO), where the absolute scale cannot be estimated from a single camera.
A visual-inertial navigation system~(VINS) enables precise mapping and localization by combining proprioceptive information from the IMU and the visual information of the surroundings from camera(s)~\cite{song2022dynavins,rovio, okvis,campos2021orb,qin2018vins}.

\begin{figure}[!t]
  \centering
  \captionsetup{font=footnotesize}
  \includegraphics[width=0.95\columnwidth]{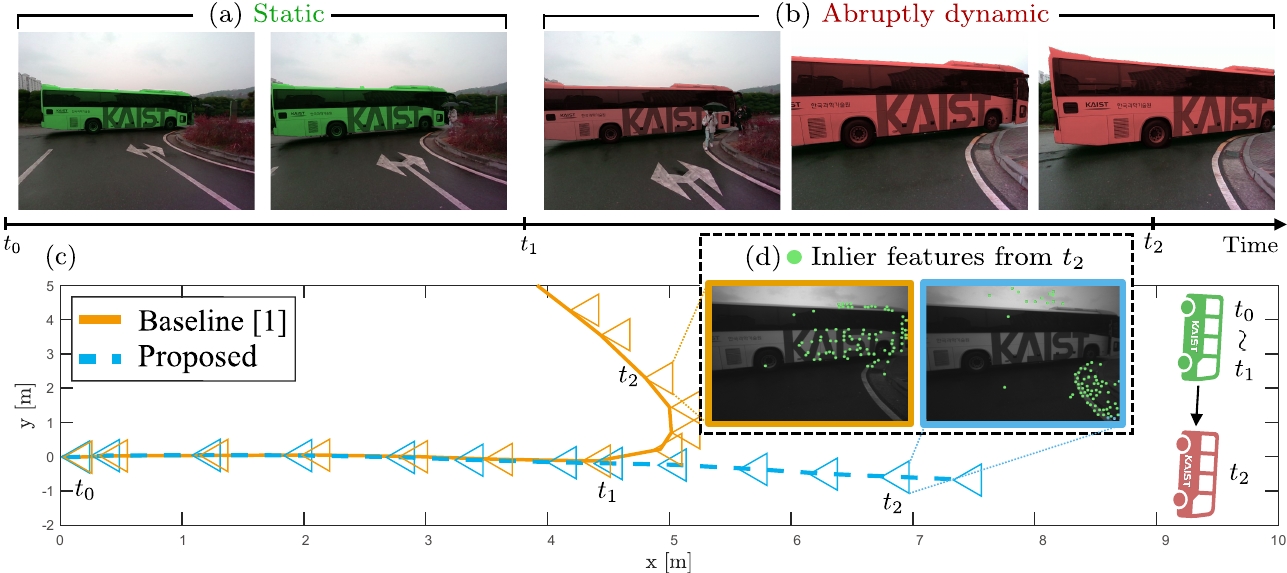}\\[-0.6ex]
  \caption
  {
  	Scenes with an abruptly dynamic object in the real world.
  	(a)~The bus and its features were initially static.
  	(b)~The bus moved suddenly during observation.
  	(c)~The estimated trajectory for each method in this case. 
  	(d)~The baseline method assumes that the features from the abruptly dynamic object are static,
  	leading to divergence in the opposite direction of the bus's movement.
  	In contrast, the proposed method effectively reduces the effect of the abruptly dynamic object
  	and uses the static features as inliers, resulting in an accurate trajectory estimate.
  }
  \label{fig:mainfigure}
  \vspace{-4mm}
\end{figure}

While VIO usually works well in static environments, its performance may be potentially degraded when several moving objects exist in the surroundings, because features originating from moving objects are not repeatedly observed in the same position; thus, using these features as correspondences results in imprecise pose estimation. 

To solve the problem, numerous researchers~\cite{fan2019dynamic,palazzolo2019refusion,dai2020rgb} have proposed robust methods for dynamic scenes by checking whether visual features are static or not, via geometry-based verification.
However, these methods assume that the number of static features remains greater than that of dynamic features.
However, this assumption does not hold when moving objects occupy most of the regions on the image plane, thus significantly increasing the number of dynamic features.

As the deep learning era has come, deep learning-based dynamic object rejection methods~\cite{vins-dynamic,bescos2018dynaslam}
have been introduced by combining the multiple-view geometry and semantic segmentation methods.
For instance, Henein~\textit{et al.}~\cite{henein2020dynamic} proposed a semantic \ac{SLAM} method by investigating a geometrical consistency across multiple views to reduce the influence of imprecise semantic masks.
Moreover, Bescos~\textit{et al.}~\cite{bescos2021dynaslam} integrated the velocity terms of the moving objects into the \ac{BA}, which allowed to perform moving object tracking and pose estimation simultaneously.

However, these deep learning-based methods only work on a few predefined dynamic classes, such as cars or humans.
Therefore, these approaches present two potential limitations: a)~the system fails to reject the influence of dynamic objects not included in the predefined classes, and b)~if an object categorized as dynamic classes remains stationary, e.g. a parked car, its static features may be wrongly rejected.
To tackle this problem of dependency on predefined labels, Blum~\textit{et al.}~\cite{SCIM} have proposed a method that can perceive unlabeled objects by checking the inaccurately mapped labels in the volumetric mapping; however, the method cannot distinguish between static and dynamic objects.
He \textit{et al.}~\cite{he2021using} proposed an extended VINS-Mono that tracks features within the masks of predefined objects by masking them before the RANSAC process. However, they assumed that outliers outside the masks could be filtered out solely through RANSAC, which might not be sufficient.

Meanwhile, outlier rejection methods have been introduced to reject the impact of outliers in optimization.
Loxham \textit{et al.}~\cite{loxam2008student} proposed a robust, real-time visual tracking filter based on a mixture of Student-\textit{t} modes for all distributions of measurements.
Switchable constraints~\cite{switchableconstraints} and dynamic covariance scaling~\cite{dcs} were introduced to eliminate false-positive loop closures during the pose graph optimization process.
Lee \textit{et al.}~\cite{lee2014solution} proposed a method to overcome the limitations of switchable constraints~\cite{switchableconstraints} and dynamic covariance scaling~\cite{dcs}.
Yang \textit{et al.}~\cite{yang2020graduated} implemented the \ac{B-R duality}~\cite{black1996unification} in various fields, from pose graph optimization to point cloud registration.

In our previous work~\cite{song2022dynavins}, we implemented a VIO framework with \ac{B-R duality} and succeeded in the rejection of false correspondences from dynamic objects. 
However, the method was limited by its slow computation speed in consequence of the additional optimization procedures. 
Furthermore, two main limitations on dynamic object rejection remained.

First, a divergence occurs when errors from these false correspondences incorrectly propagate into the states, particularly affecting the IMU bias terms.
In particular, these errors in visual tracking cause the states to converge to an incorrect local optimum.
Therefore, it triggers rapid changes in the bias estimates despite the modeling assumes that bias changes little over time~\cite{DeepIMUBias}.
Consequently, this propagated error in the bias causes the divergence of the velocity and pose estimation in the VINS.
To solve the IMU bias problem, Buchanan \textit{et al.}~\cite{DeepIMUBias} proposed a learning-based bias correction approach; however, this approach requires a pre-training phase for each IMU device.
Moreover, this method does not address the false correspondences caused by dynamic objects, which results in the recurrence of divergence.

Second, the objects that change their state from being stationary to motion, which are referred to as \textit{abruptly dynamic objects}, were not handled.
For example, as illustrated in \figref{fig:mainfigure}, if a bus is once parked and then starts to move while being observed, the speed and status of it gradually change; thus, it becomes almost unable to strictly classify whether this bus is a dynamic object or not. Accordingly, the correspondences from the abruptly dynamic object are difficult to be rejected by prior motion from IMU preintegration~\cite{song2022dynavins}.
These abruptly dynamic objects can exacerbate the aforementioned divergence issue~\cite{kundu2011realtime}.

\begin{figure}[t]
  \centering
  \captionsetup{font=footnotesize}
  {\includegraphics[width=0.95\columnwidth]{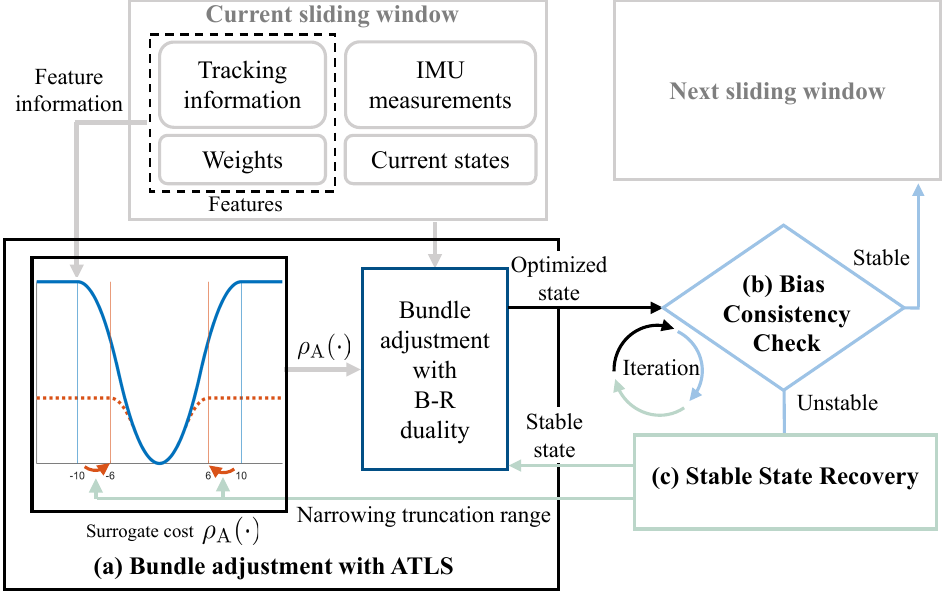}}\\[-0.2ex]
  \caption
  {
    Pipeline of the proposed robust VIO framework.
    (a) The truncation range for the surrogate cost, a form of truncated least squares, is defined by feature information.
    Initial range is shown by the blue vertical lines.
    Concurrently, the bundle adjustment (BA) process is optimized using Black-Rangarajan (B-R) duality.
    (b) The stability of the optimized result is checked using bias.
    (c) If the optimized states are determined to be unstable, all states are recovered to the previous stable values, and the truncation range of the surrogate cost is narrowed as shown by the orange vertical lines.
    Otherwise, the process moves to the next sliding window, and the states are stored.
  }
  \label{fig:outline}
  \vspace{-4mm}
\end{figure}

In this study, as illustrated in \figref{fig:outline}, we introduce a robust VIO approach, called \textit{DynaVINS++}, based on adaptive truncated least squares~(ATLS) to overcome the effects of moving objects.
Furthermore, we propose bias consistency check~(BCC) and stable state recovery~(SSR) designed to prevent divergence even after being affected by abruptly dynamic objects.

Our contributions are summarized as follows:
\begin{itemize}
	\item  
	A robust and fast VIO method is proposed to overcome the effects of dynamic objects in \ac{BA} by rejecting the residuals from such objects using truncated least squares as a surrogate cost function.
	\item
	A novel range adjustment of truncated least squares based on the current state is proposed to handle the inaccuracy of IMU preintegration owing to the aggressiveness of motion.
	\item 
	BCC and SSR are proposed to prevent divergence by checking abnormally updated IMU terms owing to abruptly dynamic objects and to revert the state to the previously updated values, respectively.
\end{itemize}

The remainder of this letter is organized as follows.
\secref{sec:methodA} discusses the limitations of the conventional VINS and proposes robust optimization using ATLS.
In \secref{sec:methodB}, we discuss the bias related problems in state estimation and propose the BCC and SSR to solve the problems.
In \secref{sec:result}, we compare our proposed method with other state-of-the-art methods in various environments.

	\makeatletter
\def\maketag@@@#1{\hbox{\m@th\normalfont\normalsize#1}}
\makeatother

\section{Robust Optimization using Adaptive Truncated Least Squares}
\label{sec:methodA}

\subsection{Notation}
\label{sec:method_notation}
Three coordinates are used in this letter. $(\cdot)^{w}$, $(\cdot)^{b}$, and $(\cdot)^{c}$ represent the world, body, and camera coordinates, respectively. 
The relative coordinate of $B$ with respect to $A$ is expressed as $(\cdot)^{A}_{B}$.
In conventional VIO methods, the $k$-th state $\mathbf{x}_k$ in the sliding window contains the position $\mathbf{p}^w_{b_k}\in\mathbb{R}^3$, velocity $\mathbf{v}^w_{b_k}\in\mathbb{R}^3$, unit-quaternion rotation $\mathbf{q}^w_{b_k}$, accelerometer bias $\mathbf{b}_{a_k}\in\mathbb{R}^3$, and gyroscope bias $\mathbf{b}_{w_k}\in\mathbb{R}^3$. 
The state vectors used in our algorithm are as follows:
\begin{equation} \label{eq:states}
	\begin{split}
		\mathcal{X} = 		   [&\mathbf{x}_1,\mathbf{x}_2,\cdots,\mathbf{x}_{n_k},\mathcal{F}],\\
		\mathbf{x}_k = 		   [&\mathbf{p}^w_{b_k},\mathbf{v}^w_{b_k},\mathbf{q}^w_{b_k},\mathbf{b}_{a_{k}},\mathbf{b}_{w_{k}}],  1 \leq k \leq n_k,\\
		\mathbf{f}_j=		   [&\lambda_j,\omega_j,\mathcal{C}_j] \in \mathcal{F},\\
 	\end{split}
\end{equation}
where $\mathcal{F}$, $\mathcal{X}$, and  $n_k$  represent the entire features, state in the sliding window, and the number of keyframes in the current sliding window, respectively.

Furthermore, the feature information $\mathbf{f}_j$ includes the inverse depth $\lambda_j$ as well as the weight $\omega_j \in [0,1]$ that represents a higher probability of being static as it approaches to one.
That is, a robust feature that is more likely to be static has a weight close to one~\cite{song2022dynavins}.
A group of camera frames in which $\mathbf{f}_j$ is tracked is denoted as
$\mathcal{C}_j,$ which contains the camera frame~$c_k$ corresponding to the state $\mathbf{x}_k$.




\subsection{Conventional Visual Inertial State Estimator}
\label{sec:method_conventionalVINS}

In the conventional visual-inertial state estimator~\cite{qin2018vins}, the formulation of \ac{BA} is represented as follows:
\begin{equation}
	\begin{gathered}
		\min_{\mathcal{X}} \left\{
		\parallel \mathbf{r}_p - \mathbf{H}_p \mathcal{X} \parallel ^2 
		+ \sum_{\hat{\mathbf{z}}^{b_k}_{b_{k+1}}\in \mathcal{I}}\parallel{\mathbf{r}_{\mathcal{I}}
			(\hat{\mathbf{z}}^{b_k}_{b_{k+1}}, \mathcal{X}) \parallel ^2_{\mathbf{P}^{b_k}_{b_{k+1}}}} 
		\right. \\ \left.
		+ \sum_{\mathbf{f}_j \in \mathcal{F}_v} \sum_{\hat{\mathbf{z}}^{c_k}_{j} \in\mathcal{V}_j} \rho_H\Big(\parallel{\mathbf{r}_{\mathcal{V}}
			(\hat{\mathbf{z}}^{c_k}_{j}, \mathcal{X}) \parallel }^2_{\mathbf{P}^{c_k}_{j}} \Big)
		\right\},
		\label{eq:convention_ba}
	\end{gathered}
\end{equation}
\noindent where $\mathcal{X}$ is a state defined in \eqref{eq:states}; $\rho_H(\cdot)$ denotes the Huber loss~\cite{huber1992robust}; $\mathbf{r}_p$, $\mathbf{r}_{\mathcal{I}}(\cdot, \cdot)$, and $\mathbf{r}_{\mathcal{V}}(\cdot, \cdot)$ represent the residuals for marginalization, IMU, and visual reprojection measurements, respectively; $\mathbf{H}_p$ denotes a measurement estimation matrix of the marginalization; $\mathbf{P}$ denotes the covariance of each term; and $\mathcal{F}_v \subset \mathcal{F}$ denotes the sufficiently tracked features, which means that insufficiently tracked features in $\mathcal{F}$ are not used in optimization.
$\mathcal{I}$ and $\mathcal{V}_j$ denote the set of the observations from the IMU preintegration and $j$-th feature, respectively.
For convenience, $\parallel\mathbf{r}_{\mathcal{I}}(\hat{\mathbf{z}}^{b_k}_{b_{k+1}}, \mathcal{X}) \parallel ^2_{\mathbf{P}^{b_k}_{b_{k+1}}}$ and $\parallel{\mathbf{r}_{\mathcal{V}}(\hat{\mathbf{z}}^{c_k}_{j}, \mathcal{X}) \parallel }_{\mathbf{P}^{c_k}_{j}}$ will be simplified as $\mathcal{R}_{\mathcal{I}}(\hat{\mathbf{z}}^{b_k}_{b_{k+1}})$ and $\mathcal{R}_{\mathcal{V}}(\hat{\mathbf{z}}^{c_k}_{j})$ later on.
As a result of the BA, estimated state $\hat{\mathcal{X}}$ can be obtained.

Once some false correspondences occur as moving objects occupy most of the regions of the image plane, the performance of the conventional visual-inertial state estimator is potentially degraded.
This is because these false correspondences from moving objects cause erroneous visual residuals, which corresponds to the third summand in~\eqref{eq:convention_ba}, making the estimated state converge into an incorrect optimum.

To solve this problem, we propose ATLS that rejects these false correspondences by adaptively adjusting the truncation range depending on the maximum residual of the tracked features~(see \secref{sec:method_ANC}).

\subsection{Weighted Parallax for Robust Keyframe Selection}
\label{sec:method_WP}

Before introducing ATLS, we propose a novel keyframe selection method based on weighted parallax, complementing the existing parallax-based method~\cite{qin2018vins} to select keyframes that are robust to moving objects.
In static environments, the keyframe is determined if the average parallax of the tracked features between the current frame and the latest keyframe exceeds a certain threshold, implying that the viewpoint difference between two frames is sufficiently large~\cite{qin2018vins}.
However, when a moving object is observed on the image plane, the average parallax is affected by the tracked features from the moving objects. 
Consequently, two primary issues may arise: First, a keyframe with an insufficiently large actual viewpoint difference may be selected, leading to inaccurate feature triangulation. Second, the keyframe may not be selected even if the viewpoint difference is sufficiently large, leading to the loss of feature tracking.

To solve this issue, the weighted average parallax $\theta_{\text{avg}}$ of the current frame is proposed using the parallax of the continuously tracked features as follows:

\begin{equation} \label{eq:avg_para}
	\theta_{\text{avg}} = \Big({\sum_{\mathcal{F}_c}\omega_j\theta_j}\Big) \Big/ \Big({\sum_{\mathcal{F}_c}\omega_j}\Big),
\end{equation}
where $\omega_j$ denotes the weight of the $j$-th feature as defined in \secref{sec:method_notation}, and will be estimated as in \secref{sec:method_ANC}. $\mathcal{F}_c$ represents the tracked features and $\theta_j$ denotes the parallax corresponding to $\mathbf{f}_j \in \mathcal{F}_c$.
If all the weights are zeros, indicating that large occlusion occurs, the sliding window is reset and the next frame is set as the first keyframe.

\begin{figure}[t]
  \centering
  \captionsetup{font=footnotesize}
  
  \subfigure[]{\includegraphics[width=0.85\columnwidth]{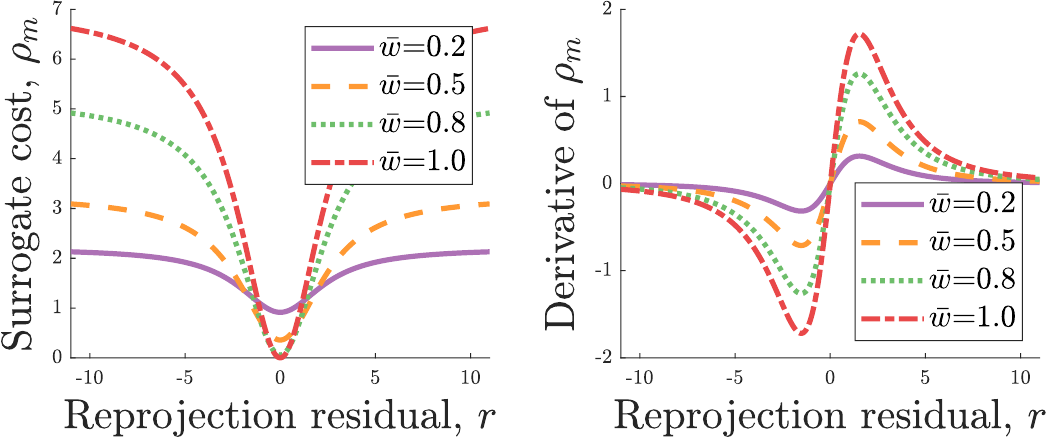}}\\[-0.2ex]
  \subfigure[]{\includegraphics[width=0.85\columnwidth]{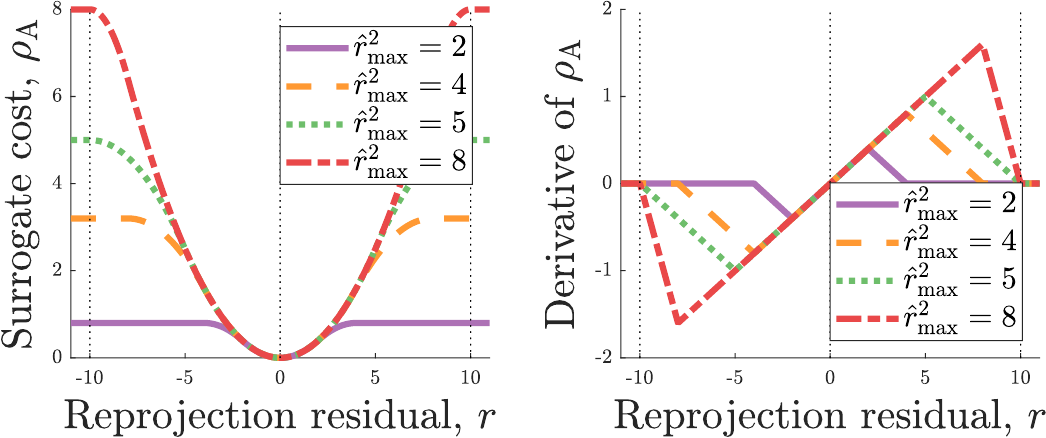}}\\[-1.8ex]
  \caption
  {
	(a)~Surrogate cost and the corresponding derivative (gradients) with respect to the previous weight $\bar{w}$ in DynaVINS~\cite{song2022dynavins}.
	It is shown that the gradient cannot be zero for all ranges.
	(b)~Surrogate cost and the gradients with respect to the maximum residual of static features, $\hat{r}^2_{\text{max}}$, in the current frame in our proposed method.
	The gradient is zero when the residual exceeds the predefined maximum threshold, 10 in this figure, or $2\hat{r}_{\text{max}}^2$.
  }
  \label{fig:tls}
  \vspace{-4mm}
\end{figure}

\subsection{Adaptive Truncated Least Squares (ATLS)}
\label{sec:method_ANC}

As described in DynaVINS~\cite{song2022dynavins}, existing non-linear kernels, such as Huber loss, \ac{TLS}, and Geman-McClure are insufficient to fully reject the effect of dynamic features.
Therefore, DynaVINS adopted alternating optimization by employing a surrogate cost, which is designed in the form of TLS to eliminate the influence of features with a residual over a certain threshold.

However, there still exist two problems.
First, as shown in \figref{fig:tls}(a), the gradient of the surrogate cost for a large residual in DynaVINS is near zero, but not exactly zero.
Consequently, the impact of outliers is not completely rejected during the \ac{BA}, leading to imprecise pose estimation.
Second, additional computations are required to unnecessarily update the depths and weights of these outlier correspondences.

To solve these problems, we propose ATLS, a method that employs a surrogate cost function $\rho_\mathrm{A}(\cdot)$ to completely truncate the impact of dynamic features by setting the gradient for the large residual as zero, as shown in \figref{fig:tls}(b).
Therefore, our approach effectively minimizes the impact of outliers while reducing unnecessary computational burden~(see \secref{sec:result_ablation}).
Furthermore, our ATLS adjusts the truncation range based on the maximum residual of tracked features to better distinguish between features from static objects and those from dynamic objects in the current scene.

Formally, by applying weights and the penalty term to the third summand instead of the Huber loss $\rho_H$ in \eqref{eq:convention_ba}, our objective function is defined in the form of \ac{B-R duality} to get the estimated state $\hat{\mathcal{X}}$ as follows:
\begin{equation}
\begin{gathered}
\hat{\mathcal{X}} = \argmin_{\mathcal{X}} \left\{
\parallel \mathbf{r}_p - \mathbf{H}_p \mathcal{X} \parallel ^2 
+ \sum_{\hat{\mathbf{z}}^{b_k}_{b_{k+1}}\in \mathcal{I}} \mathcal{R}_{\mathcal{I}}(\hat{\mathbf{z}}^{b_k}_{b_{k+1}})^2
\right. \\ \left.
+ \sum_{\mathbf{f}_j \in \mathcal{F}_v} 
	\Big\{ \omega_j (\sum_{\hat{\mathbf{z}}^{c_k}_{j} \in\mathcal{V}_j} 
	\mathcal{R}_{\mathcal{V}}(\hat{\mathbf{z}}^{c_k}_{j})^2)+\Phi_{\rho_\mathrm{A}}(\omega_j)\Big\}
\right\},
\label{eq:anc_ba}
\end{gathered}
\end{equation}
where $\Phi_{\rho_\mathrm{A}}(\cdot)$ is a penalty term corresponding to $\rho_\mathrm{A}(\cdot)$.

To solve~\eqref{eq:anc_ba}, we adopt the alternating optimization~\cite{black1996unification} by alternately updating the weights and the penalty term as follows.

\subsubsection{Weight Update} \label{sec:method_ANC_weight}

The weight update involves a)~setting the adaptive truncation range and b)~assigning the weight using the residual of each correspondence.
We observe that the range of the residuals from static features in the current frame is not always constant because the residuals are affected by the current pose $\mathbf{x}_{n}$ derived from the IMU preintegration.
Thus, we adjust the shape of the surrogate cost using the estimated static features whose $\omega_j = 1$, and their current residuals based on $\mathbf{x}_{n}$. Thus, the adjusted surrogate cost reduces the wrong rejection of previous static features even though aggressive body motion occurs.

To adaptively determine the range of \ac{TLS}, we use two terms: the user-defined global maximum residual threshold,~$r_\text{max}$, and the maximum residuals of the tracked features that are considered as static, $\hat{r}_\text{max} \geq 0$.
Note that $r_\text{max}$ should be larger than $\hat{r}_\text{max}$. Then, the truncation range $r_\text{trunc}$ can be defined as follows:
\begin{equation} \label{eq:settrunc}
	r_\text{trunc} = \min \{r_\text{max}, \: 2 \hat{r}_{\text{max}}\}.
\end{equation}

Here, we describe the method for selecting the estimated static features that have been tracked for a long time to set $\hat{r}_{\text{max}}$.
Originally, the features used in optimization, $\mathcal{F}_v$, can be divided into two groups: The tracked features, $\mathcal{F}_c$, which was previously used in~\eqref{eq:avg_para}, and the features that were tracked within the sliding window but are no longer tracked in the current frame, $\mathcal{F}_l$, where $\mathcal{F}_c \cap \mathcal{F}_l = \varnothing$.
Then, $\mathcal{F}_c$ is further categorized into two feature sets: features that have been optimized,~$\mathcal{F}_o$, and features that have not yet been optimized owing to insufficient tracking duration,~$\mathcal{F}_n$.

By denoting the residual of $\mathbf{f}_j \in \mathcal{F}_o$ by $\hat{r}_j = \mathcal{R}_{\mathcal{V}}(\hat{\mathbf{z}}^{c_n}_{j})$, the residuals from $\mathcal{F}_o$ whose weights are one are used to set $\hat{r}_{\text{max}}$ as follows:
\begin{equation}
	\hat{r}_{\text{max}} = \sqrt{ \text{max}\Big\{\hat{r}_j^2 \mid \mathbf{f}_j \in \mathcal{F}_o \land \omega_j=1\Big\}}.
\end{equation}

Next, to assign weights for the surrogate cost $\rho_{\text{A}}(\cdot)$ and compute the penalty term $\Phi_{\rho_\mathrm{A}}(\cdot)$, the following equation should be satisfied according to \cite{yang2020graduated}:
\begin{equation} \label{eq:howto}
	\rho_{\text{A}}(\hat{r}_j^2) = \omega_j \hat{r}_j^2 + \Phi_{\rho_\text{A}}(\omega_j).
\end{equation}

The penalty term can be derived using a procedure provided by Black and Rangarajan~\cite{black1996unification} to make the surrogate cost $\rho_\mathrm{A}(\cdot)$ in a TLS form so that the ranges $r_\text{trunc}$ and $\hat{r}_{\text{max}}$ as shown in \figref{fig:tls}(b) satisfy the followings:
\begin{equation} \label{eq:surrogate}
	\Phi_{\rho_\mathrm{A}}(\omega_j) = \mu \hat{r}_{\text{max}} r_\text{trunc} \Big(\frac{1-\omega_j}{\mu+\omega_j}\Big),
\end{equation}
where $\mu = \hat{r}_{\text{max}} / (r_\text{trunc} - \hat{r}_{\text{max}})$.
Note that when $\hat{r}_\text{max} < 0.5 r_\text{max}$, $\mu$ is set to one according to~\eqref{eq:settrunc}.
This implies that all the residuals of the static features are less than half of the predefined global maximum residual, $r_\text{max}$, in the current frame, indicating that the IMU and visual information are well-matched.

The weight update can be solved in a closed form as follows:
\begin{equation}
	\label{eq:weight_update}
	\hat{\omega}_j =
	\begin{cases}
		0	&	\text{if} \quad \hat{r}^2_j \in [{r}^2_{\text{trunc}},+\infty)\\
		\mu({r}_{\text{trunc}}/\hat{r}_j - 1/\hat{r}_{\text{max}}) &\text{if} \quad \hat{r}^2_j \in [\hat{r}^2_{\text{max}},{r}^2_{\text{trunc}})\\
		1	&	\text{if} \quad \hat{r}^2_j \in [0,\hat{r}^2_{\text{max}}).
	\end{cases}
\end{equation}
Finally, weights are updated as $\omega_j \leftarrow \min(\hat{\omega_j},\omega_j)$, to prevent misclassification of dynamic features as static in the current frame.
As a result, the obtained weights can act as $\rho_{A}(\cdot)$ in state optimization, as shown in the blue solid curve at the bottom left of \figref{fig:outline}.

Note that, for $\mathcal{F}_l$, the weights are not updated because these features are no longer observed in the current frame.
Consequently, it is assumed that these features have been sufficiently optimized in previous frames.
For $\mathcal{F}_n$, the value of $\hat{r}_j$ in~\eqref{eq:weight_update} is determined as the maximum reprojection error across all observations within the window.
Specifically, $\hat{r}_j$ is defined as~\mbox{$\hat{r}_j = \text{max} \Big\{\mathcal{R}_{\mathcal{V}}(\hat{\mathbf{z}}^{c_k}_{j}) \mid \mathbf{z}^{c_k}_{j} \in\mathcal{V}_j\Big\}$}, thereby conservatively accounting for potential inaccuracies.


\subsubsection{State Optimization}
After the weights of features are updated, the states are optimized using a non-minimal solver~\cite{agarwal2015others} with the fixed weights as follows:
\begin{equation}
	\begin{gathered}
		\hat{\mathcal{X}} = \argmin_{\mathcal{X}} \left\{
		\parallel \mathbf{r}_p - \mathbf{H}_p \mathcal{X} \parallel ^2 
		+ \sum_{\hat{\mathbf{z}}^{b_k}_{b_{k+1}}\in \mathcal{I}}\{\mathcal{R}_{\mathcal{I}}(\hat{\mathbf{z}}^{b_k}_{b_{k+1}})^2\}
		\right. \\ \left.
		+ \sum_{\mathbf{f}_j \in \mathcal{F}_v} 
		\{ \omega_j\sum_{\hat{\mathbf{z}}^{c_k}_{j} \in\mathcal{V}_j} 
		\mathcal{R}_{\mathcal{V}}(\hat{\mathbf{z}}^{c_k}_{j})^2\}
		\right\},
		\label{eq:anc_ba_state}
	\end{gathered}
\end{equation}
\noindent which can be considered a weighted BA formulation compared with~\eqref{eq:convention_ba}.
Note that the penalty term in~\eqref{eq:anc_ba} is discarded because it is constant when the weights are fixed.
	\makeatletter
\def\maketag@@@#1{\hbox{\m@th\normalfont\normalsize#1}}
\makeatother

\section{Addressing Abruptly Dynamic Objects in Terms of IMU Biases} \label{sec:methodB}
The method proposed in \secref{sec:methodA} effectively rejects features from dynamic objects; however, a limitation arises when ATLS incorrectly assigns the weights of features from moving objects closer to one.
This issue is particularly prevalent in cases involving abruptly dynamic objects. Such misclassification can produce incorrect calculations of the truncation range in \eqref{eq:settrunc}, potentially resulting in inaccurately estimated states.

To address the aforementioned limitation, a method for enhancing the robustness of both pose and IMU bias estimation is proposed in this section.

\subsection{IMU Biases in The State Estimation}
\label{sec:effect_bias}
Before introducing our BCC and SSR, we briefly explain how IMU bias terms are used in state estimation.
As presented in \cite{qin2018vins}, the IMU measurement residual $\mathbf{r}_{\mathcal{I}} (\hat{\mathbf{z}}^{b_k}_{b_{k+1}}, \mathcal{X})$ is defined as follows:
{\small
\begin{equation}\label{eq:resid_IMU}
	\begin{bmatrix}
		\mathbf{R}^{b_k}_w (\mathbf{p}^w_{b_{k+1}} - \mathbf{p}^w_{b_{k}} + \frac{1}{2}\mathbf{g}^w \Delta t_k^2 - \mathbf{v}^w_{b_k}\Delta t_k) - \hat{\boldsymbol{\alpha}}^{b_k}_{b_{k+1}}\\
		\mathbf{R}^{b_k}_w (\mathbf{v}^w_{b_{k+1}}+\mathbf{g}^2 \Delta t_k - \mathbf{v}^w_{b_k}) - \hat{\boldsymbol{\beta}}^{b_k}_{b_{k+1}}\\
		2[{\mathbf{q}^w_{b_k}}^{-1} \otimes \mathbf{q}^w_{b_{k+1}} \otimes (\hat{\boldsymbol{\gamma}}^{b_k}_{b_{k+1}})^{-1}]_{xyz}\\
		\mathbf{b}_{a_{k+1}} - \mathbf{b}_{a_{k}} \\
		\mathbf{b}_{w_{k+1}} - \mathbf{b}_{w_{k}}
	\end{bmatrix},
\end{equation}}

\noindent where $\mathbf{R}$ represents a rotation matrix corresponding to quaternion~$\mathbf{q}$.
$\hat{\boldsymbol{\alpha}}^{b_k}_{b_{k+1}}$, $\hat{\boldsymbol{\beta}}^{b_k}_{b_{k+1}}$, and $\hat{\boldsymbol{\gamma}}^{b_k}_{b_{k+1}}$ denote the preintegrated IMU measurement terms between two body frames $b_k$ and $b_{k+1}$, which are recursively updated as follows:

{
\begin{equation} \label{eq:preint_repropa}
	\begin{split}
		\hat{\boldsymbol{\alpha}}^{b_k}_{b_{k+1}} &\leftarrow \hat{\boldsymbol{\alpha}}^{b_k}_{b_{k+1}} + \mathbf{J}^\alpha_{b_a} \delta \mathbf{b}_{a_k} + \mathbf{J}^\alpha_{b_w} \delta \mathbf{b}_{w_k}, \\
		\hat{\boldsymbol{\beta}}^{b_k}_{b_{k+1}}  &\leftarrow \hat{\boldsymbol{\beta}}^{b_k}_{b_{k+1}} + \mathbf{J}^\beta_{b_a} \delta \mathbf{b}_{a_k} + \mathbf{J}^\beta_{b_w} \delta \mathbf{b}_{w_k}, \\
		\hat{\boldsymbol{\gamma}}^{b_k}_{b_{k+1}} &\leftarrow \hat{\boldsymbol{\gamma}}^{b_k}_{b_{k+1}} \otimes  \begin{bmatrix} 1 \\ \frac{1}{2} \mathbf{J}^\gamma_{b_w}\delta\mathbf{b}_{w_k} \end{bmatrix},\\
	\end{split}
\end{equation}}

\noindent where $\mathbf{J}^a_b$ represents the Jacobian of $a \in \{ \boldsymbol{\alpha}^{b_k}_{b_{k+1}}, \boldsymbol{\beta}^{b_k}_{b_{k+1}} \}$ with respect to $b \in \{ \mathbf{b}_{a_k}, \mathbf{b}_{w_k} \}$.
Therefore, after the IMU preintegration, $\hat{\boldsymbol{\alpha}}$, $\hat{\boldsymbol{\beta}}$, and $\hat{\boldsymbol{\gamma}}$ can be adjusted only by the IMU bias because Jacobians are constant.

For brevity, we will express the top three residual terms of \eqref{eq:resid_IMU} as $\mathbf{r}_{\mathcal{I}} (\cdot, \cdot)_{\boldsymbol{\alpha}\boldsymbol{\beta}\boldsymbol{\gamma}}$ later on. 

\subsection{Effect of Abruptly Dynamic Objects on IMU Biases}
\label{sec:effect_abruptly}

While estimating the current state through the framework presented in \secref{sec:methodA}, the weights of features from abruptly dynamic objects can be set to one.
This is because the abruptly dynamic objects are sufficiently tracked over time when they are static, i.e.~the weights of the features from these objects are incorrectly assigned to a value of $\omega=1$, even when they have a large $\hat{r}_j$, resulting in the setting of an incorrect truncation range in~\eqref{eq:settrunc}.
Consequently, once $r_\text{trunc}$ is increased, indicating that the non-convexity of the surrogate cost is relaxed, ATLS no longer effectively rejects the influence of features from moving objects.

\begin{figure}[t]
  \centering
  \captionsetup{font=footnotesize}
  \includegraphics[width=0.9\columnwidth]{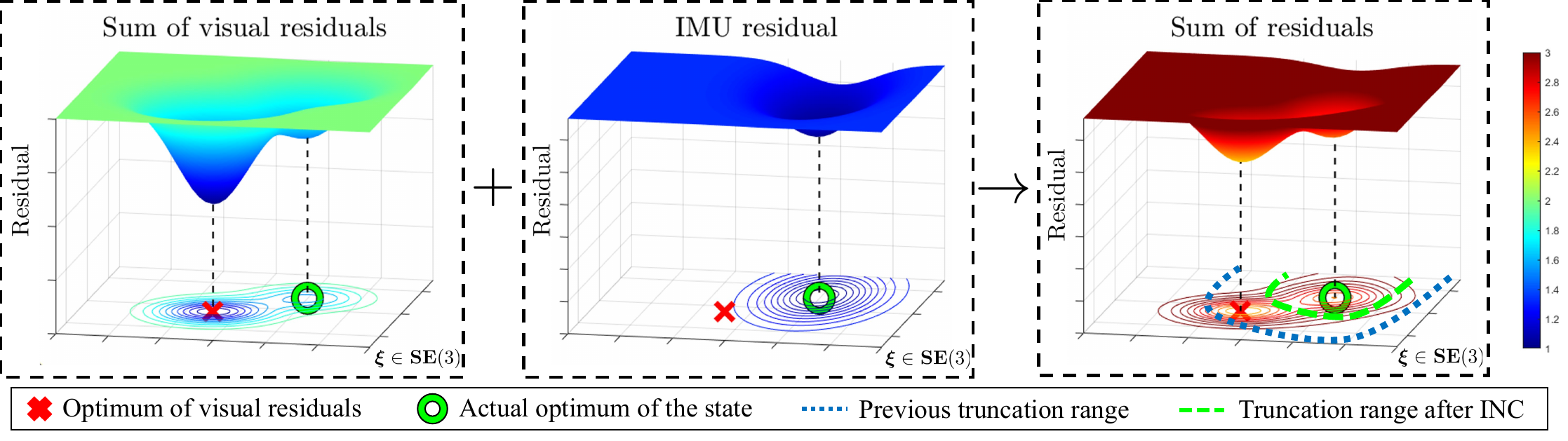}\\[-0.8ex]
  \caption
  {
    Visual description of the residual for the current frame when an abruptly dynamic object exists~($\mathbf{SE}(3)$ is projected onto the 2D for better visualization).
    The current state can fall into incorrect optimum if the influence of the visual residual becomes greater than the IMU residual.
    By increasing the non-convexity of the surrogate cost~(from the blue dotted line to the green dashed line), the optimizer can ignore the visual residual and make the state converge into the actual optimum.
  }
  \label{fig:wrongopt}
  \vspace{-6mm}
\end{figure}

In particular, as shown in \figref{fig:wrongopt}, when features from abruptly dynamic objects are included in the BA, the discrepancy between the optimum of visual residuals~(red $\times$ mark) and the actual optimum of the state~(green $\circ$ mark) occurs, triggering wrong state estimation.
We observed that the state diverges when the errors from these false correspondences propagate into the IMU bias terms in the sliding window.

Therefore, to prevent divergence, we propose BCC to check if the bias terms are misaligned with the visual information affected by the abruptly dynamic object rather than the actual body motion.
If the misalignment is detected, SSR is applied to revert the states to their previous values.
Subsequently, we re-optimize them within a narrower truncation residual range for the optimizer to make the state converge into the actual optimum, as shown in \figref{fig:wrongopt}. 

\subsection{Bias Consistency Check}
\label{sec:method_BSC}

The motivation of BCC is that when an abruptly dynamic object starts to move within the sliding window,
the estimated pose and bias terms of $\mathbf{x}_k$ in the sliding window may become misaligned to each other owing to false correspondences from the abruptly dynamic object, which is referred to as the pose and bias being \textit{inconsistent} with each other.

\newcommand{\observedframe}{l}
In particular, by denoting the frame where the abruptly dynamic object is observed as the $\observedframe$-th frame in the sliding window,
the incorrect optimum of the state owing to the abruptly dynamic object makes the poses after the $\observedframe$-th frame erroneous after BA,
while the poses prior to the $\observedframe$-th frame are relatively unaffected because most of their correspondences are from static features.
In contrast to the pose terms, the bias terms after the $\observedframe$-th frame, which also become erroneous owing to the incorrect optimum,
directly affect the bias terms prior to the $\observedframe$-th frame because they are modeled to have the same values over time, i.e.~$\mathbf{b}_{a_{k+1}} - \mathbf{b}_{a_{k}} = 0$ and $\mathbf{b}_{w_{k+1}} - \mathbf{b}_{w_{k}} = 0$, as presented in \eqref{eq:resid_IMU}.
Consequently, this inconsistency between pose and bias terms prior to the $l$-th frame results in a significant difference in the values of $\mathbf{r}_{\mathcal{I}} (\cdot, \cdot)_{\boldsymbol{\alpha}\boldsymbol{\beta}\boldsymbol{\gamma}}$ before and the after optimization.

Therefore, we propose BCC to detect the inconsistency by comparing the $\text{L}_{\text{2}}$ norm of $\mathbf{r}_{\mathcal{I}} (\cdot, \cdot)_{\boldsymbol{\alpha}\boldsymbol{\beta}\boldsymbol{\gamma}}$ with optimized bias and pose terms against the $\text{L}_{\text{2}}$ norm with the bias terms before the optimization.

Let $\hat{\mathcal{X}}^-$ consist of the bias terms before the optimization and the optimized poses,
we examine the consistency by checking the number of inconsistent frames within the sliding window, $n_{\text{a}}$, which is defined as follows:
\begin{equation}\label{eq:n_a}
	n_{\text{a}} = \text{card}\bigg( k \,\bigg|\, \frac
	{\parallel \mathbf{r}_{\mathcal{I}} (\hat{\mathbf{z}}^{b_k}_{b_{k+1}}, \hat{\mathcal{X}})_{\boldsymbol{\alpha}\boldsymbol{\beta}\boldsymbol{\gamma}} \parallel}
	{\parallel \mathbf{r}_{\mathcal{I}} (\hat{\mathbf{z}}^{b_k}_{b_{k+1}}, \hat{\mathcal{X}}^{-})_{\boldsymbol{\alpha}\boldsymbol{\beta}\boldsymbol{\gamma}} \parallel}
	> \tau_r \,\bigg) , \, k < n_k-1,
\end{equation}
where $\tau_r \geq 1$ denotes the residual ratio threshold, and card($\cdot$) represents the cardinality, i.e. the number of elements.
Therefore, if $n_{\text{a}}$ is over a certain threshold $\tau_a$ after the BA, it implies that the optimized current state was affected by abruptly dynamic objects, and thus, current bias terms in ${\hat{\mathcal{X}}}$ are inaccurate.

\subsection{Stable State Recovery and Re-Optimization}
\label{sec:method_BRINC}

If the inconsistency is not detected, the updated state $\hat{\mathcal{X}}$ is confirmed.
Otherwise, we revert the states to their values before the optimization, in the SSR module.
Then, we re-optimize the states with a modified kernel to prevent the states from converging to an incorrect optimum.
To this end, $r_\text{trunc}$ is divided by 2 to reduce the range widened in \eqref{eq:settrunc}.
Then, the weights are updated by \eqref{eq:weight_update} and the optimization process is conducted again with updated weights.

Therefore, our proposed method iterates the alternating optimization performed in ATLS, as well as the entire optimization procedure, as illustrated in \figref{fig:outline}.


	\section{Experimental Results And Discussion}
\label{sec:result}
We evaluated the effectiveness of our proposed approach in terms of performance and computational speed.
Additionally, we compared our algorithm with the following state-of-the-art~(SOTA) algorithms: ORB-SLAM3~\cite{campos2021orb}, DynaSLAM~\cite{bescos2018dynaslam}, and our previous work, DynaVINS~\cite{song2022dynavins}, across various scenes.
We tested both a mono-inertial (\texttt{-M-I}) and a stereo-inertial~(\texttt{-S-I}) mode for each algorithm. All tests were performed on a Ryzen 3900x CPU.
For our method, we set $\tau_r = \tau_a = 2$, $n_k = 9$, $n_o = 4$, and $r_\text{max} = 10$ pixels.

\subsection{Datasets and Error Metric}
\label{sec:result_dataset}
\subsubsection{VIODE Dataset} 
The VIODE dataset~\cite{minoda2021viode} is a simulated dataset that includes moving objects, such as cars or trucks,
making it suitable for evaluating the effect of dynamic objects on VIO methods. 
The dataset also contains large occlusion, where most of the image is covered by dynamic objects.
The number of dynamic objects in the scene is denoted by the sub-sequence name \texttt{none} to \texttt{high}.

\subsubsection{Our Dataset}  
To observe the effect of abruptly dynamic objects, we acquired two additional sequences, namely \texttt{Lateral} and \texttt{Parallel}, to evaluate the impact of such objects on VIO methods. 
In the \texttt{Lateral} and \texttt{Parallel} sequences, the abruptly dynamic object moves laterally and parallelly to the robot's direction of movement, respectively.

\subsubsection{Error Metric}
Absolute trajectory error (ATE)~\cite{zhang2018tutorial} was used to evaluate the accuracy of the estimated trajectory in this study.
To evaluate the impact of the abruptly dynamic object on state estimation accuracy over time, the relative translation error (RTE) was used.

\subsection{Ablation Study on Proposed Modules}
\label{sec:result_ablation}
To validate the improvement in accuracy and computation time of the proposed modules, we evaluated the combination of these modules as well as VINS-Fusion and DynaVINS, as shown in Table~\ref{table:ablation}.
For a fair comparison, VINS-Fusion, DynaVINS, and our method used identical parameters: stereo-inertial(\texttt{-S-I}) mode, a maximum of 200 features, and a minimum distance between features of 15 pixels.
Remarkably, as demonstrated in Table~\ref{table:ablation}, our approach using only ATLS showed substantial performance increase compared with VINS-Fusion, without an increase in time consumption.
Furthermore, the ATE is even lower when our method is augmented with the other proposed modules.

\begin{table}[t]
	\centering
	\abovecaptionskip
	\belowcaptionskip
	\renewcommand{\arraystretch}{1.2}
	
	\captionsetup{font=footnotesize}
	\caption{Ablation experiment to validate the enhancement in the accuracy and computation time of each module in \texttt{parking\_lot high} sequence.}
	\scriptsize
	\begin{tabular}{ccc|c|c|c}
		\hline\hline
			\multicolumn{3}{c|}{\multirow{2}{*}{Method}} & 
			\multirow{2}{*}{ATE [m]} & 
			\multicolumn{2}{c}{Computation time [ms]} \\
			\cline{5-6}
			 & & & 
			 & \multirow{1}*{\Gape[-2pt]{\makecell{BA}}}
			 & \multirow{1}*{\Gape[-2pt]{\makecell{Marginalization}}}\\
			\hline 
			\multicolumn{3}{c|}{{VINS-Fusion~\cite{qin2018vins}}}    	
			& 0.300     		& 36.6348  & \textbf{6.7007}  \\
			\multicolumn{3}{c|}{{DynaVINS~\cite{song2022dynavins}}}    
			&0.148          	& 74.5828  & 18.0461 \\
			\hline
			\multicolumn{3}{c|}{Proposed method} & & & \\
			ATLS & BCC & SSR &  & & \\
			\cdashline{1-3}
			\checkmark	& 	  &   	 & 0.162 & \textbf{35.9103} & 8.9690  \\
			\checkmark	& \checkmark &    & 0.112 & 40.1704 & 9.6655  \\
			\checkmark	& \checkmark & \checkmark & \textbf{0.091} & 43.7155 & 9.5851  \\
		\hline\hline
	\end{tabular}
	\label{table:ablation}
	\vspace{-6mm}
\end{table}

\begin{table*}[t]

	\centering
	\captionsetup{font=footnotesize}
	\caption{Comparison with state-of-the-art methods in VIODE~\cite{minoda2021viode} dataset (RMSE of ATE in [m]).}
    \scriptsize
    \abovecaptionskip
    \belowcaptionskip
    \setlength{\aboverulesep}{0pt}
    \setlength{\belowrulesep}{0pt}
    \renewcommand{\arraystretch}{1.2}
	\begin{tabular}{c|cccc|cccc|cccc}
	\hline\hline
	\multirow{2}{*}{Method} & \multicolumn{4}{c|}{\texttt{city\_day}} & \multicolumn{4}{c|}{\texttt{city\_night}} & \multicolumn{4}{c}{\texttt{parking\_lot}}\\
	\cline{2-13}
	&\texttt{none} & \texttt{low} & \texttt{mid} & \texttt{high} & \texttt{none} & \texttt{low} & \texttt{mid} & \texttt{high} & \texttt{none} & \texttt{low} & \texttt{mid} & \texttt{high} \\
	
	\midrule 
	{ORB-SLAM3\texttt{-M-I}~\cite{campos2021orb}} & 1.940 & 0.857 & 4.486 & * & * & * & * & * & 0.147 & 0.175 & 0.145 & 0.194 
	 \\  
	{VINS-Fusion\texttt{-M-I}~\cite{qin2018vins}} & \underline{0.210} & 0.182 & 0.560 & 0.510 & 0.328 & 0.371 & 0.457 & 0.464 & 0.102 & 0.138 & 0.707 & 1.135 
	\\ 
	DynaVINS\texttt{-M-I}~\cite{song2022dynavins}
	& 0.224 & \underline{0.167} & \textbf{{0.154}} & \underline{0.364} 
	& \underline{0.189}	& \textbf{{0.181}} & \underline{0.184} & \underline{0.256} 
	& \underline{0.097} & \underline{0.120} & \underline{0.118} & \underline{0.149}
	 \\
	
	\textbf{Ours\texttt{-M-I}} 
	& \textbf{0.162} & \textbf{0.162} & \underline{0.158} & \textbf{0.150}
	& \textbf{0.187} & \underline{0.193} & \textbf{0.172} & \textbf{0.249}
	& \textbf{0.088} & \textbf{0.110} & \textbf{0.115} & \textbf{0.122}
	\\ 
	
	\midrule
	DynaSLAM\texttt{-S}~\cite{bescos2018dynaslam}
	&  1.621 & 1.426 & 1.638 & * 
	& 3.333 & 3.314 & 3.074 & 3.865 
	&  0.108 & 0.170 & * & * 
	\\ 
	 
	ORB-SLAM3\texttt{-S-I}~\cite{campos2021orb}
	& 0.302 & 0.419 & 0.217 & * 
	& 0.709 & 0.895 & 1.693 & 3.006
	& 0.148 & \textbf{0.067} & * & * 
	 \\ 
	 
	VINS-Fusion\texttt{-S-I}~\cite{qin2018vins}
	& \underline{0.150} & 0.203 &  0.234 & 0.373 
	& 0.317 & 0.507 & 0.494 & 0.828 
	& 0.121 & 0.121 & 0.212 & 0.300
	\\ 
	 
	DynaVINS\texttt{-S-I}~\cite{song2022dynavins}
	& 0.171 & \underline{0.178} & \underline{0.206} & \underline{0.285}
	& \underline{0.266} & \underline{0.295} & \underline{0.321} & \textbf{0.275}
	& \textbf{0.079} & 0.123 & \underline{0.095} & \underline{0.148}
    \\	
	
	\textbf{Ours\texttt{-S-I}} 
	& \textbf{0.149} & \textbf{0.129} & \textbf{0.136} & \textbf{0.168}
	& \textbf{0.252} &\textbf{0.294} & \textbf{0.235} & \underline{0.277}
	& \underline{0.106} &  \underline{0.091} &  \textbf{0.083} &  \textbf{0.091}
	\\	
	\hline\hline
	\end{tabular}
	\vspace{-1mm}
	\label{table:sota_comparison}
     \begin{flushleft}
      \footnotesize *: Failure (diverged), \texttt{-M}: Mono, \texttt{-S}: Stereo, and \texttt{-I}: Inertial. The \textbf{bold} and \underline{underlined} texts stand for the best and second best accuracy, respectively.    \end{flushleft}
      \vspace{-6mm}
\end{table*}
\begin{figure}[t]
  \centering
  \captionsetup{font=footnotesize}
  \subfigure[]{\includegraphics[width=0.45\columnwidth]{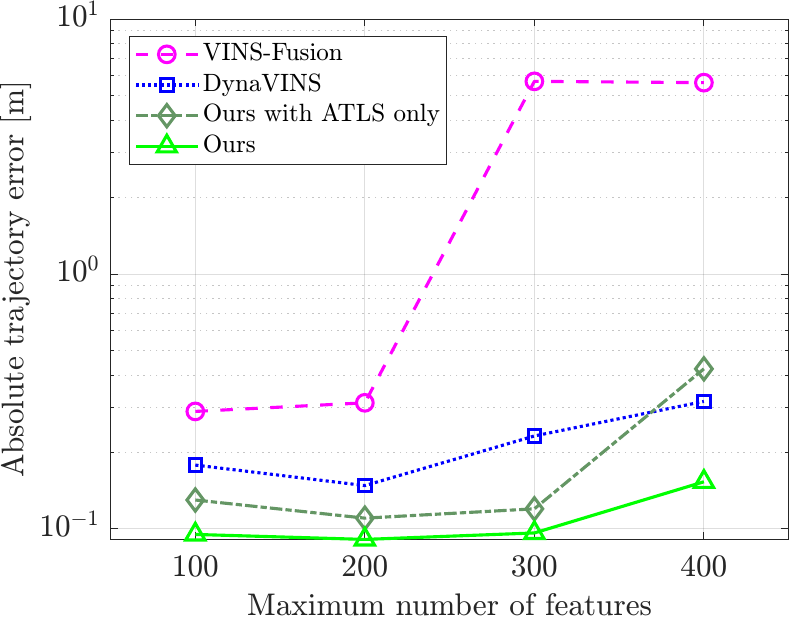}}
  \subfigure[]{\includegraphics[width=0.455\columnwidth]{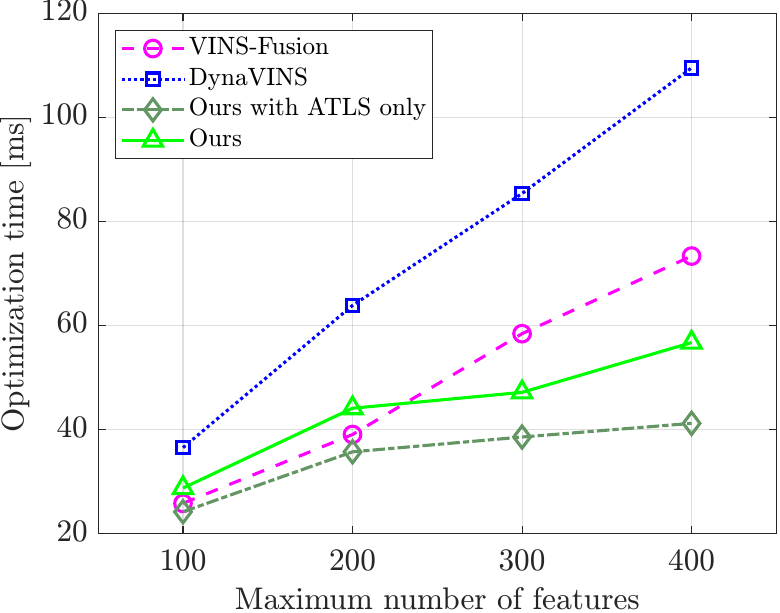}}\\[-1ex]
  \caption
  {
  (a) Absolute trajectory error and (b) the average BA time with respect to the maximum number of features in \texttt{parking\_lot} \texttt{high} sequence.
  }
  \label{fig:timecomp_ablation}
  \vspace{-5mm}
\end{figure}

\begin{figure}[!ht]
  \centering
  \captionsetup{font=footnotesize}
  \subfigure[]{\includegraphics[width=0.8\columnwidth]{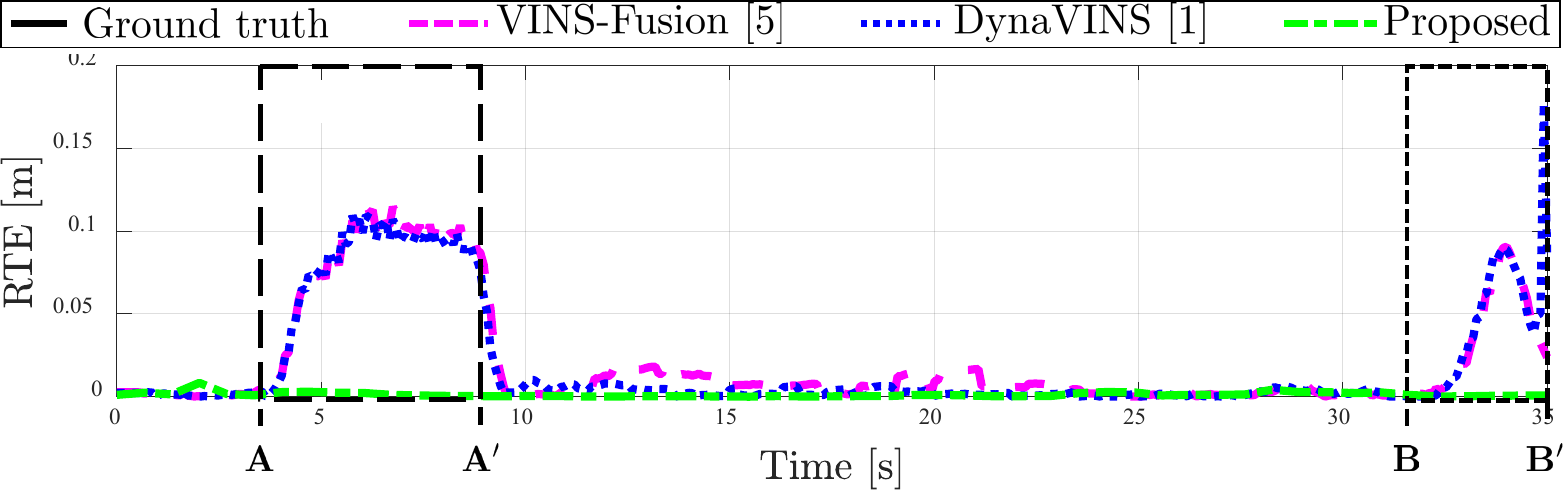}}\\[-0.7ex]
  \subfigure[]{\includegraphics[width=0.8\columnwidth]{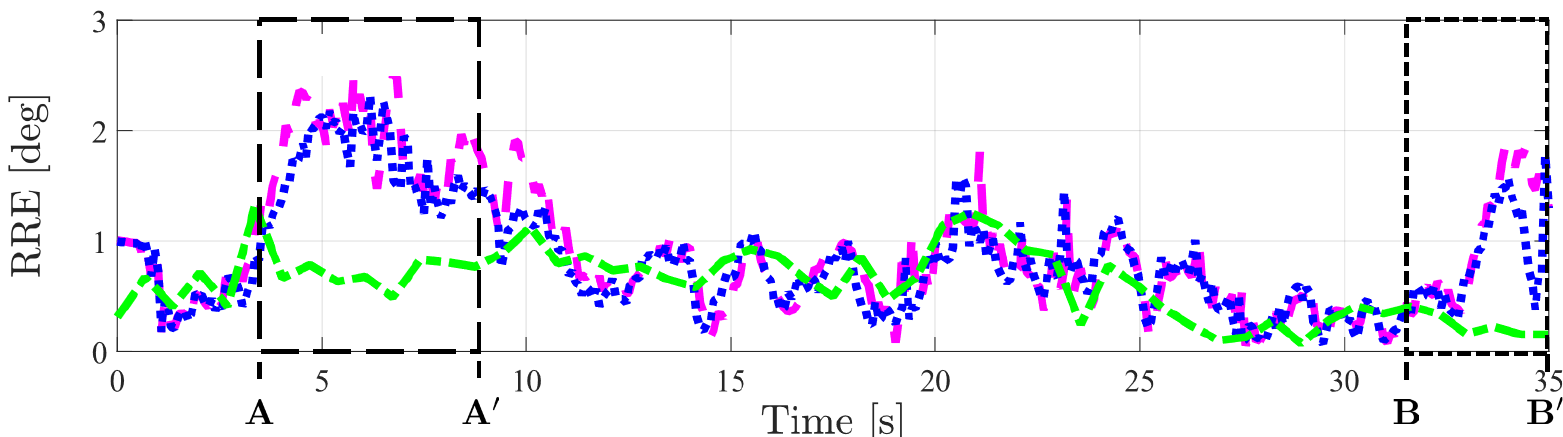}}\\[-0.7ex]
  \subfigure[]{\includegraphics[width=0.9\columnwidth]{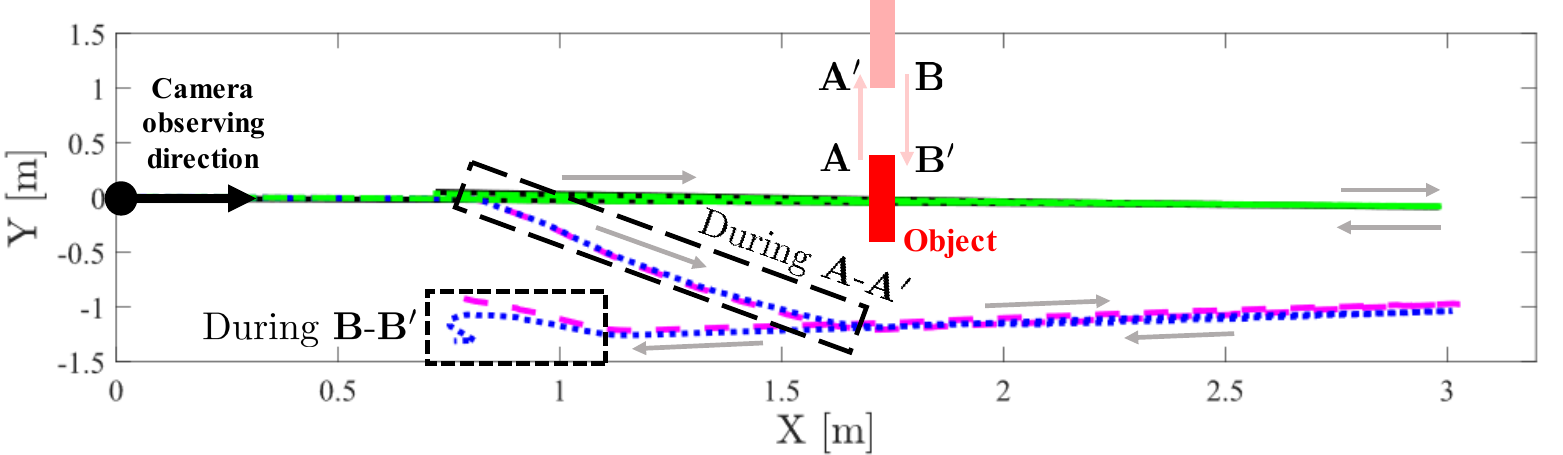}}\\[-0.8ex]
  \subfigure[]{\includegraphics[width=0.9\columnwidth]{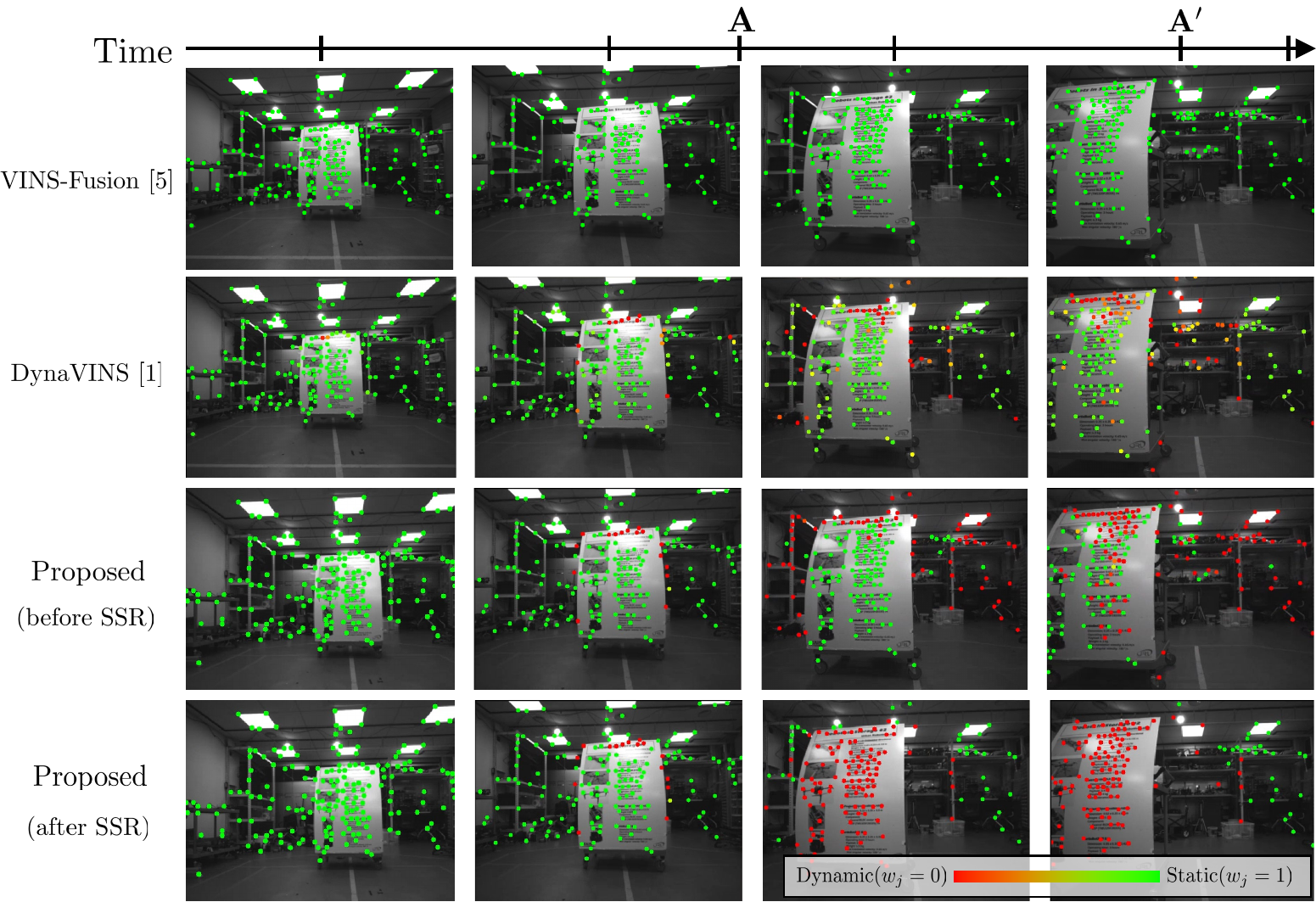}}\\[-1ex]
  \caption
  {   	
  	(a)--(b) The relative translation error (RTE) and relative rotation error (RRE) per 0.2 m of each algorithm over time.
  	An object suddenly moves at times \textbf{A} and \textbf{B}.
  	(c) Estimated trajectory from each algorithm in the \texttt{Lateral} sequence. 
  	When the abruptly dynamic object moves suddenly, the trajectory from our proposed method aligns closely with the GT trajectory. 
  	In contrast, the trajectories from other two algorithms diverge in the opposite direction of the object's movement.
  	(d)	Feature weight ($\omega$) results when the object moves at time \textbf{A}. 
  	VINS-Fusion used all tracked features for BA, wrongly considering features from abruptly dynamic objects as static.
  	DynaVINS wrongly classified features from the abruptly dynamic object as static, even after its movement. 
  	Furthermore, features in static environments were classified as dynamic.
  	Our approach, when implemented without SSR, showed results similar to those from DynaVINS.
  	However, with SSR, our approach successfully rejected the features on the object and utilized static features in the environment.
  }
  \label{fig:abrupttest1}
  \vspace{-10mm}
\end{figure}

\begin{figure}[!ht]
	\centering
	\captionsetup{font=footnotesize}
	\subfigure[]{\includegraphics[width=0.8\columnwidth]{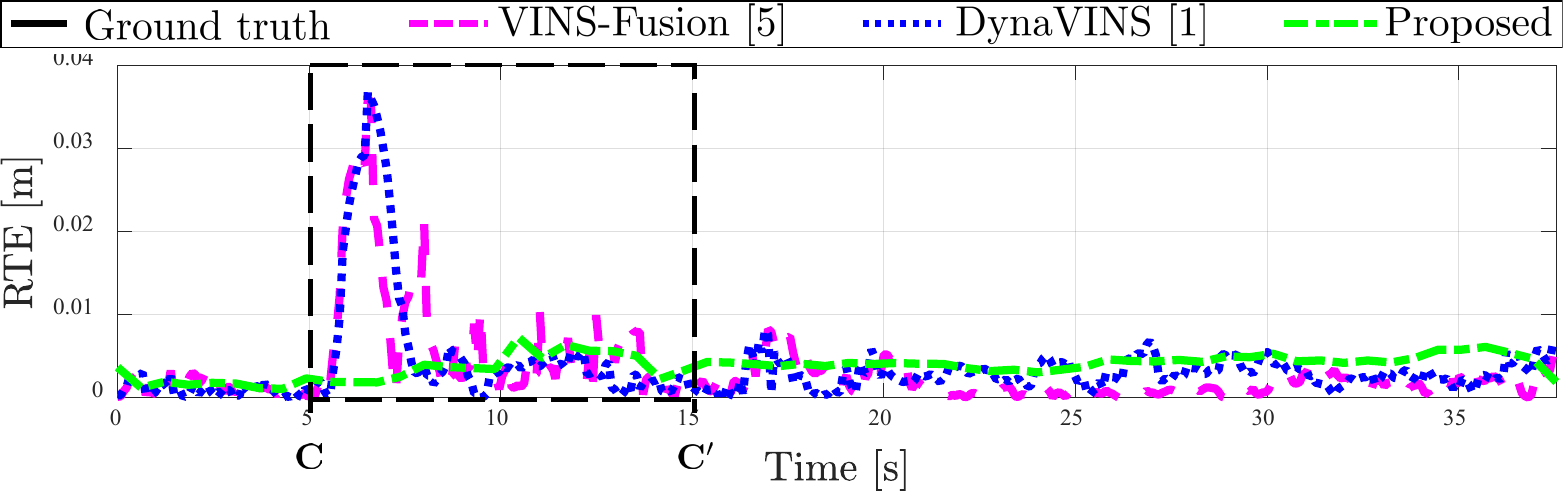}}\\[-1ex]
	\subfigure[]{\includegraphics[width=0.8\columnwidth]{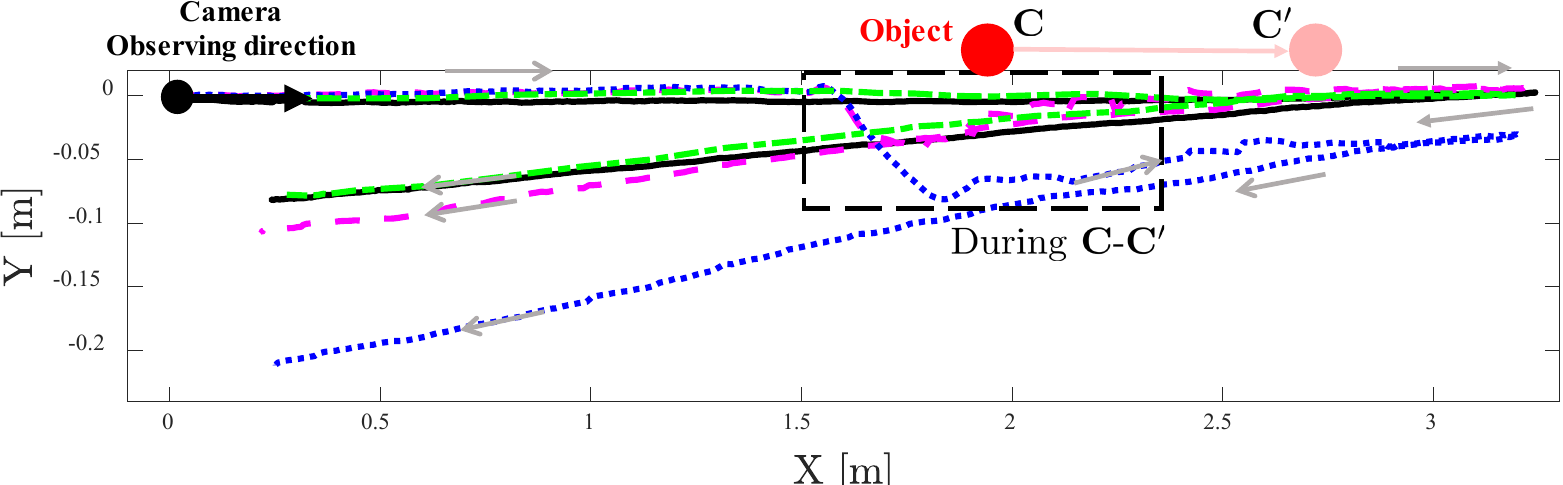}}\\[-2ex]
	\caption
	{   	
		(a) RTE per 0.2m of each algorithm over time. When the object suddenly moves at time \textbf{C}, the RTE of other algorithms increases; while ours remains accurate.
		(a) Estimated trajectories of algorithms in the \texttt{Parallel} sequence. 
		When the abruptly dynamic object moves suddenly at time \textbf{C}, trajectories from VINS-Fusion and DynaVINS struggled. 
	}
	\label{fig:abrupttest2}
\vspace{-5mm}
\end{figure}

The second evaluation aimed to analyze the accuracy and computation time regarding the maximum number of features.
As the number of total features increases, the number of dynamic features also increases and their residuals can influence more on the BA.
As shown in \figref{fig:timecomp_ablation}(a), DynaVINS and our approach showed smaller error than VINS-Fusion, while VINS-Fusion exhibited degraded accuracy with the abused features.
It is because the correspondences from moving objects were successfully rejected even if there are excessive number of features in our method.
Furthermore, our method did not increase the computation time significantly compared with VINS-Fusion as the number of tracked features increases, whereas the computation time of DynaVINS proportionally increased, as depicted in \figref{fig:timecomp_ablation}(b).
This is because our method reduces the number of features used in BA by rejecting dynamic features.
This result suggests that our approach can be applied across various environments with varying numbers of features, without the risk of encountering detrimental effects from dynamic objects when using excessive number of features.

\subsection{Comparison With Other Algorithms}
\label{sec:result_tevaluation}
\subsubsection{Robustness to Consistently Dynamic Objects}
As presented in \tabref{table:sota_comparison}, the SOTA algorithms estimate accurate poses in static environments.
However, as the number of dynamic objects increased (sequences from \texttt{none} to \texttt{high}), the performances of VINS-Fusion and ORB-SLAM3 were substantially degraded.
DynaSLAM was less accurate than others, even compared to the mono-inertial case of VINS-Fusion, DynaVINS, and ours, because stereo-only with a short baseline is worse at estimating metrics than the IMU. 
Moreover, the trajectories of DynaSLAM in \texttt{parking\_lot} sequences diverged because DynaSLAM could not detect objects when they are too close to the camera.

In situations with overall occlusion in the \texttt{parking\_lot high} and \texttt{city\_day high} sequences, VINS-Fusion tried to estimate the poses through IMU preintegration, even when there were few features to track. 
Nevertheless, VINS-Fusion showed higher errors in these sequences than those  in \texttt{low} sequences.
Similarly, owing to the large occlusion in the image plane by moving objects,
DynaVINS also exhibited low accuracy in these scenarios because the algorithm could not recover a stable state.

In our proposed algorithm, even the impact of dynamic objects was not perfectly removed only by the ATLS module, the BCC module found the inconsistency and SSR module restored the stable state to avoid the impacts from the dynamic objects.
Consequently, our algorithm showed better performance in most sequences compared with others.

\subsubsection{Robustness Against Abruptly Dynamic Objects}
The robustness of algorithms is compared in the case of the existence of abruptly dynamic objects using our dataset.
In our dataset, DynaSLAM and ORB-SLAM3 diverged and we could not represent the path results.

In the \texttt{Lateral} sequence, an object (poster board) which was static suddenly moves to the left side of the image plane at time \textbf{A} while the robot is moving forward, as shown in Figs.~6(c)--(d). 
VINS-Fusion estimated the body poses in the same direction with the object's movement as it considers the object to be static.
As shown in the second row of \figref{fig:abrupttest1}(d), DynaVINS could not swiftly change weights of the features owing to its momentum factors, leading to a failure in rejecting features from the abruptly dynamic object. 
Moreover, new static features were wrongly considered as being dynamic owing to the dominant effect of the abruptly dynamic object.
Consequently, VINS-Fusion and DynaVINS had higher errors when the object moved, as shown in Figs.~6(a) and (b) for translation and rotation errors, respectively.
The same problem occurred when our BCC and SSR are not employed as shown in the third row of \figref{fig:abrupttest1}(d).
However, when the bias inconsistency was detected, the stable state was recovered, and the truncation range was more conservatively adjusted to reject features from abruptly dynamic objects, as shown in the forth row of \figref{fig:abrupttest1}(d). Then, our algorithm could resume the robust state estimation and maintain low error.

Similarly, in the \texttt{Parallel} sequence, while other algorithms showed an inaccurate trajectory owing to the effect of the abruptly dynamic object, ours demonstrated more precise trajectory, as shown in \figref{fig:abrupttest2}. This is because the ATLS, BCC, and SRR in our proposed method enable robust state estimation against the abruptly moving object.

These results demonstrate that our proposed ATLS, BCC, and SSR modules can successfully reduce the influence of abruptly dynamic objects.

	\section{Conclusions}
\label{sec:concolusion}

In this study, we have proposed a VIO framework robust against dynamic objects, particularly abruptly dynamic objects.
By employing ATLS, our method can reject the various dynamic objects while maintaining real-time performance.
Moreover, we demonstrated that our BCC and SSR could successfully address the bias problem caused by abruptly dynamic objects.
In further studies, we plan to suggest a framework to determine whether an unlabeled object is static or dynamic, by combining our method with SCIM~\cite{SCIM}.
	
	\renewcommand*{\bibfont}{\small}
	\bibliographystyle{IEEEtranN} 
	\bibliography{string-short,references}

	\vfill
	
\end{document}